\def\year{2021}\relax
\documentclass[letterpaper]{article} 
\usepackage{aaai21}  
\usepackage{times}  
\usepackage{helvet} 
\usepackage{courier}  
\usepackage[hyphens]{url}  
\usepackage{graphicx} 
\usepackage{natbib}  
\usepackage{caption} 

\usepackage{amsmath}
\usepackage{amsfonts}
\usepackage[ruled,vlined]{algorithm2e}
\usepackage{graphicx}
\usepackage{multirow}
\usepackage{lipsum}
\usepackage{bbding}
\usepackage{placeins}

\title{A Joint Training Dual-MRC Framework for Aspect Based Sentiment Analysis}
\author{

    Yue Mao, Yi Shen, Chao Yu, Longjun Cai
    \\
}
\affiliations{

    Alibaba Group, Beijing, China\\

    \{maoyue.my, sy133447, aiqi.yc, longjun.clj\}@alibaba-inc.com

}

\begin{document}

\maketitle

\begin{abstract}
Aspect based sentiment analysis (ABSA) involves three fundamental subtasks: aspect term extraction, opinion term extraction, and aspect-level sentiment classification. 
Early works only focused on solving one of these subtasks individually. 
Some recent work focused on solving a combination of two subtasks, e.g., extracting aspect terms along with sentiment polarities or extracting the aspect and opinion terms pair-wisely. 
More recently, the triple extraction task has been proposed, i.e., extracting the (aspect term, opinion term, sentiment polarity) triples from a sentence. 
However, previous approaches fail to solve all subtasks in a unified end-to-end framework.
In this paper, we propose a complete solution for ABSA. 
We construct two machine reading comprehension (MRC) problems and solve all subtasks by joint training two BERT-MRC models with parameters sharing. 
We conduct experiments on these subtasks, and results on several benchmark datasets demonstrate the effectiveness of our proposed framework,  which significantly outperforms existing state-of-the-art methods.
\end{abstract}

\section{Introduction}
\begin{figure}[h]
    \centering
    \includegraphics[width=0.45\textwidth]{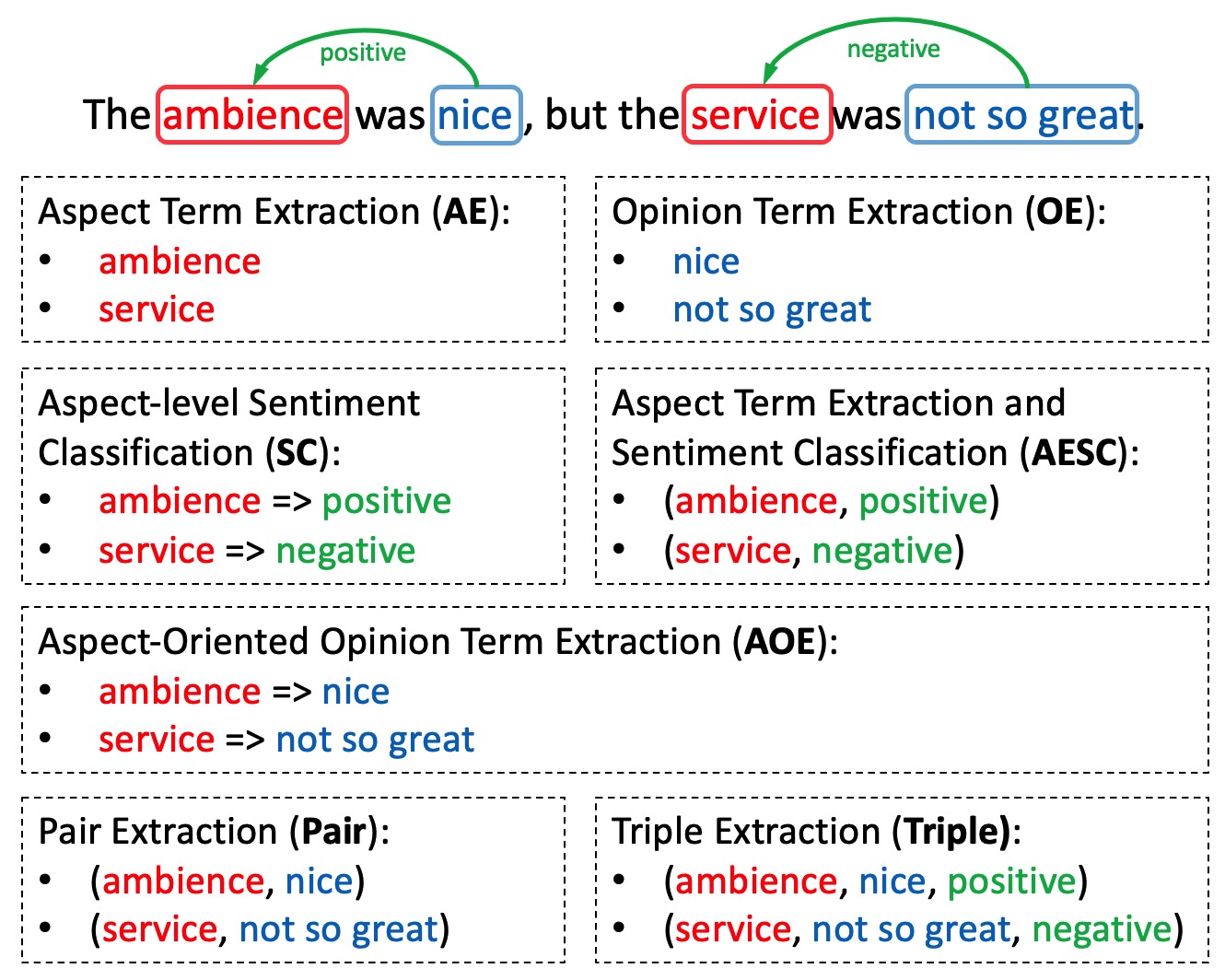} 
    \caption{An illustrative example of ABSA substasks.} \label{example}
\end{figure}

Aspect based sentiment analysis (ABSA)\footnote{It is also referred as target based sentiment analysis (TBSA).} is an important research area in natural language processing.
Consider the example in Figure \ref{example}, in the sentence ``\emph{The ambience was nice, but the service was not so great.}'', 
the aspect terms (AT) are ``\emph{ambience/service}'' and the opinion terms (OT) are ``\emph{nice/not so great}''. Traditionally, there exist three fundamental subtasks: aspect term extraction, opinion term extraction, and aspect-level sentiment classification. 
Recent research works aim to do a combination of two subtasks and have achieved great progress. For example, they extract (AT, OT) pairs, or extract ATs with corresponding sentiment polarities (SP). 
More recently, some work that aims to do all related subtasks in ABSA with a unified framework has raised increasing interests.

For convenience, we assume the following abbreviations of ABSA subtasks as illustrated in Figure \ref{example}:  
 \begin{itemize}
    \item \textbf{\emph{AE}}: AT extraction
    \item \textbf{\emph{OE}}: OT extraction
    \item \textbf{\emph{SC}}: aspect-level sentiment classification
    \item \textbf{\emph{AESC}}\footnote{It is also referred as aspect based sentiment analysis (ABSA).}: AT extraction and sentiment classification
    \item \textbf{\emph{AOE}}\footnote{It is also referred as target oriented opinion word extraction (TOWE).}: aspect-oriented OT extraction
    \item \textbf{\emph{Pair}}: (AT, OT) pair extraction 
    \item \textbf{\emph{Triple}}: (AT, OT, SP) triple extraction.
\end{itemize}

We mainly focus on the task of extracting $(a,o,s)$ triples since it is the hardest among all ABSA substasks.   
\citet{peng2020knowing} proposed a unified framework to extract (AT, OT, SP) triples. 
However, it is computationally inefficient as its framework has two stages and has to train three separate models.

In this paper, we propose a joint training framework to handle all ABSA subtasks (described in Figure \ref{subtasks}) in one single model.
We use BERT \cite{devlin2019bert} as our backbone network and use a span based model to detect the start/end positions of ATs/OTs from a sentence.
Span based methods outperform traditional sequence tagging based methods for extraction tasks \cite{hu-etal-2019-open}.    
Following its idea, a heuristic multi-span decoding algorithm is used, which is based on the non-maximum suppression algorithm (NMS) \cite{Rosenfeld:1971:ECD}.

We convert the original triple extraction task to two machine reading comprehension (MRC) problems.
MRC methods are known to be effective if a pre-trained BERT model is used. 
The reason might be that BERT is usually pre-trained with the next sentence prediction to capture the pairwise sentence relations.
Theoretically, the triple extraction task can be decomposited to subtasks \emph{AE}, \emph{AOE} and \emph{SC} .
Thus, we use the left MRC to handle \emph{AE} and the right MRC to handle \emph{AOE} and \emph{SC}. 
Our main contributions in this paper are as follows:
\begin{itemize}
    \item We show the triple extraction task can be jointly trained with three objectives.
    \item We propose a dual-MRC framework that can handle all subtasks in ABSA (as illustrated in Table \ref{subtasks}).
    \item We conduct experiments to compare our proposed framework on these tasks. Experimental results show that our proposed method outperforms the state-of-the-art methods. 
\end{itemize}
\begin{table}
    \centering
    \begin{tabular}{c|c|cc}
    \hline
    \multirow{2}{*}{Subtasks}  & Left-MRC & \multicolumn{2}{c}{Right-MRC}       \\ \cline{2-4} 
           & Extraction  & \multicolumn{1}{c|}{Classification} & Extraction \\ \hline
    AE     &$\surd$    & \multicolumn{1}{c|}{}    &      \\ \hline
    AOE     &      & \multicolumn{1}{c|}{}    &$\surd$    \\ \hline
    SC     &      & \multicolumn{1}{c|}{$\surd$}   &      \\ \hline
    AESC   &$\surd$    & \multicolumn{1}{c|}{$\surd$}    &    \\ \hline
    Pair   &$\surd$    & \multicolumn{1}{c|}{}    &$\surd$    \\ \hline
    Triple &$\surd$    & \multicolumn{1}{c|}{$\surd$}   &$\surd$    \\ \hline
    \end{tabular} 
    \caption{Our proposed dual-MRC can handle all ABSA subtasks.}
    \label{subtasks}
\end{table}

\section{Related Work}
Aspect-based sentiment analysis (ABSA) has been widely studied since it was first proposed in  \cite{kddHuL04}. In this section, we present existing works on ABSA according to related subtasks.

\textbf{SC.} Various neural models have been proposed for this task in recent years. 
The core idea of these works is to capture the intricate relationship between an aspect and its context by designing various neural architectures such as 
CNN \cite{HuangC18,LamLSB18}, RNN \cite{tang-etal-2016-effective,ZhangZV16,RuderGB16}, attention-based network \cite{MaLZW17, DuSWQLXL19, WangHZZ16,GuZHS18,YangTWXC17}, memory network\cite{TangQL16,ChenSBY17,FanGD0XW18}.
\citet{SunHQ19}  convert \emph{SC} to a BERT sentence-pair classification task, which achieves state-of-the-art results of this task. 
    
\textbf{AE.} As the pre-task of \emph{SC}, \emph{AE} aims to identify all aspect terms in a sentence  \cite{kddHuL04,pontiki-etal-2014-semeval} and is usually regarded as a sequence labeling problem  \cite{LiBLLY18,XuLSY18,HeLND17}. 
Besides, \cite{MaLWXW19} and \cite{LiCQLS20} formulated \emph{AE} as a sequence-to-sequence learning task and also achieved impressive results. 	

\textbf{AESC.} In order to make \emph{AESC} meet the needs of practical use, plenty of previous works make efforts to solve \emph{AE} and \emph{SC} simultaneously. 
Simply merging \emph{AE} and \emph{SC} in a pipeline manner will lead to an error-propagation problem \cite{MaLW18}. 
Some works \cite{li2019unified,LiBZL19} attempt to extract aspects and predicting corresponding sentiment polarities jointly through sequence tagging based on a unified tagging scheme. However, 
these approaches are inefficient due to the compositionality of candidate labels \cite{LeeKP016} and may suffer the sentiment inconsistency problem. 
\citet{ZhouHGHH19} and  \citet{hu-etal-2019-open} utilize span-based methods to conduct \emph{AE} and \emph{SC} at the span-level rather than token-level, which are able to overcome the sentiment inconsistency problem. 
It is worth noting that the information of opinion terms is under-exploited during these works. 

\textbf{OE.} Opinion term extraction (\emph{OE}) is widely employed as an auxiliary task to improve the performance of \emph{AE} \cite{YuJX19,DBLP:conf/aaai/WangPDX17,PanW18}, \emph{SC} \cite{he-etal-2019-interactive} or both of them \cite{chen-qian-2020-relation}. 
However, the extracted ATs and OTs in these works are not in pairs, as a result, they can not provide  the cause for corresponding polarity of an aspect. 	

\textbf{AOE.} The task \emph{AOE} \cite{fan2019target} has been proposed for the pair-wise aspect and opinion terms extraction in which the aspect terms are given in advance. \citet{fan2019target} design an aspect-fused sequence tagging approach for this task.   
\citet{DBLP:conf/aaai/WuZDHC20} utilize a transfer learning method that leverages latent opinions knowledge from auxiliary datasets to boost the performance of \emph{AOE}. 

\textbf{Pair.}  \citet{ZhaoHZLX20} proposed the \emph{Pair} task to extract aspect-opinion pairs from scratch, they develop a span-based multi-task framework, 
which first enumerates all the candidate spans and then construct two classifiers to identify the types of spans (i.e. aspect or opinion terms) and the relationship between spans.  	

\textbf{Triple.}  \citet{peng2020knowing} defined the triple extraction task for ABSA, which aims to extract all possible aspect terms as well as their corresponding opinion term and sentiment polarity.  
The method proposed in  \cite{peng2020knowing} is a two-stage framework, the first stage contains two separate modules, one is a unified sequence tagging model for \emph{AE} and \emph{SC}, the other is a graph convolutional neural network(GCN) for \emph{OE}. 
In the second stage, all possible aspect-opinion pairs are enumerated and a binary classifier is constructed to judge whether the aspect term and opinion term match with each other.  
The main difference between our work and  \cite{peng2020knowing} is that we regard all subtasks as a question-answering problem, and propose a unified framework based on a single model.

\section{Proposed Framework}
\subsection{Joint Training for Triple Extraction}
In this section, we focus on the triple extraction task and the other subtasks can be regarded as special cases of it.
Given a sentence $x_j$ with max-length $n$ as the input. 
Let $T_j=\{(a,o,s)\}$  be the output of annotated triples given the input sentence $x_j$, where 
$s \in \{$Positive, Neutral, Negative$\}$ and $(a,o,s)$ refers to (aspect term, opinion term and sentiment polarity). 
For the training set $\mathcal{D} = \{(x_j, T_j) \}$ , we want to maximize the likelihood 
\begin{equation}
    L(\mathcal{D}) =   \prod_{j=1}^{|\mathcal{D}|}\prod_{(a,o,s) \in T_j} P((a,o,s)|x_j).
\end{equation}
Define 
\begin{equation}
T_j|a :=\{(o,s), (a,o,s) \in T_j \}, \quad k_{j, a} := |T_j|a|.
\end{equation}
Consider the log-likelihood for $x_j$,
\begin{eqnarray}
    \ell(x_j)&=& \sum_{(a,o,s) \in T_j} \log P((a,o,s)|x_j) \nonumber \\
    &=& \sum_{a \in T_j} \sum_{(o,s) \in T_j|a}  \log P(a|x_j) + \log P((o,s)|a, x_j)\nonumber \\
    &=& \sum_{a \in T_j} \left( \sum_{(o,s) \in T_j|a}  \log P(a|x_j) \right) \nonumber\\
    &+&  \sum_{a \in T_j} \left(\sum_{(o,s) \in T_j|a}  \log P(s|a, x_j) + \log P(o|a, x_j) \right)  \nonumber\\
    \end{eqnarray}
The last equation holds because the opinion terms $o$ and the sentiment polarity $s$ are conditionally independent  
given the sentence $x_j$ and the aspect term $a$.
\footnote{Note $(x_j, a)$ has all the information needed to determine $s$. The term $o$ does not bring additional information as it can be implied by $(x_j, a)$,
therefore $P(s|x_j,a,o)=P(s|x_j,a)$}

\begin{eqnarray}
    \ell(x_j) &=&  \sum_{a \in T_j}  k_{j,a} \cdot \log P(a|x_j)\nonumber \\
&+&\sum_{a \in T_j} \left(  k_{j,a} \cdot \log P(s|a, x_j) 
     + \sum_{o \in T_j|a} \log P(o|a, x_j)\right). \nonumber\\ 
\end{eqnarray}
We sum above equation over $ x_j \in \mathcal{D}$ and normalize the both sides, then we get
the log-likelihood of the following form
\begin{eqnarray}
    \ell(\mathcal{D}) &=& \alpha \cdot \sum_{j=1}^{|\mathcal{D}|} \sum_{a \in T_j} \left(\sum_{a \in T_j} \log P(a|x_j) \right) \nonumber \\ 
    &+&  \beta \cdot \sum_{j=1}^{|\mathcal{D}|} \sum_{a \in T_j}  \log P(s|a, x_j)  \nonumber \\ 
    &+&  \gamma \cdot \sum_{j=1}^{|\mathcal{D}|} \sum_{a \in T_j}  \left( \sum_{o \in T_j|a} \log P(o|a, x_j) \right) \label{eqn_convert}
\end{eqnarray}
where $\alpha,\beta, \gamma \in [0, 1]$. The first term is repeated in order to match with the other two terms.
From (\ref{eqn_convert}), we may conclude the triple extraction task \emph{Triple} can be converted to the joint training of \emph{AE}, \emph{SC} and \emph{AOE}.

\subsection{Dual-MRC Framework}
\begin{figure*}
    \centering
    \includegraphics[width=1\textwidth]{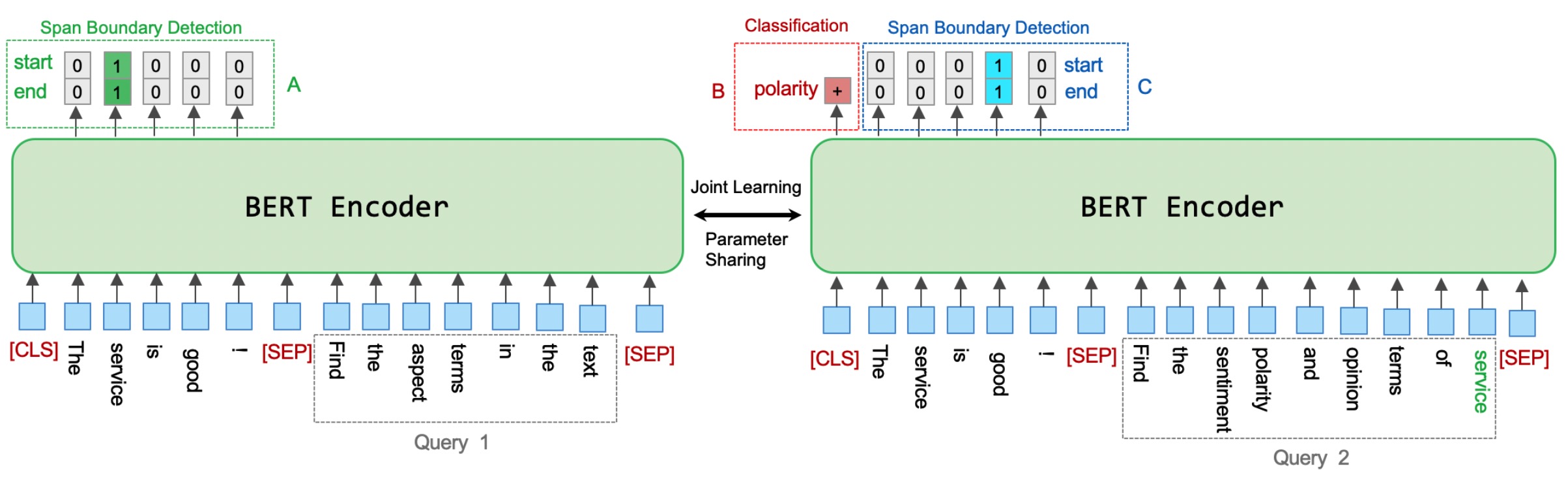}
    \caption{Proposed joint training dual-MRC framework.} \label{fig_main}
\end{figure*}
Now we are going to propose our joint training dual-MRC framework. 
As illustrated in Figure \ref{fig_main}, our model consists of two parts. 
Both parts use BERT \cite{devlin2019bert} as their backbone models to encode the context information. 
Recall that BERT is a multi-layer bidirectional Transformer based language representation model. Let $n$ denote the sentence length and $d$ denote the hidden dimension.
Suppose the last layer outputs for all tokens are $h^{l,s}, h^{r,s}, h^{l,e}, h^{r,e}\in \mathbb{R}^{(n+2) \times d}$ which are used for extraction, 
where $l/r$ refer to the left/right part and $s/e$ refer to the stard/end token. 
Suppose the output of BERT at the [CLS] token is $h^{r}_{cls} \in \mathbb{R}^{(n+2) \times d}$ which is used for classification.  

The goal of the left part is to extract all ATs from the given text, i.e., the task \emph{AE}.
As we discussed previously, span based methods are proven to be effective for extraction tasks. 
We follow the idea in \cite{hu-etal-2019-open}, for the left part, we obtain the logits and probabilities for 
start/end positions 
\begin{eqnarray}
    g^{l,s} = W^{l,s} h^{l,s},\quad p^{l, s} = softmax(g^{l,s})  \\
    g^{l,e} = W^{l,e} h^{l,e},\quad p^{l, e} = softmax(g^{l,e})
\end{eqnarray}
where $W^{l,s}\in \mathbb{R}^{1 \times d}$ and $W^{l,e}\in \mathbb{R}^{1 \times d}$ are trainable weights and softmax is taken over all tokens. 
Define the extraction loss of the left part as  
\begin{equation}
    \mathcal{J}_{AE} = - \sum_{i} y^{l,s}_i \log(p^{l,s}_i) 
- \sum_i y^{l,e}_i \log(p^{l,e}_i)
\end{equation}
where $y^{l,s} $ and $y^{l,e} $ are ground truth start and end positions for ATs.

The goal of the right part is to extract all OTs and find the sentiment polarity with respect to a given specific AT. 
Similarly, we obtain the logits and probabilities for 
start/end positions 
\begin{eqnarray}
    g^{r,s} = W^{r,s} h^{r,s},\quad p^{r, s} = softmax(g^{r,s}) \\
    g^{r,e} = W^{r,e} h^{r,e},\quad p^{r, e} = softmax(g^{r,e})
\end{eqnarray}
where $W^{r,s}\in \mathbb{R}^{1 \times d}$ and $W^{r,e}\in \mathbb{R}^{1 \times d}$ are trainable weights and softmax is applied on all tokens. 
Define the extraction loss of the right part as  
\begin{equation}\mathcal{J}_{AOE} = - \sum_{i} y^{r,s}_i \log(p^{r,s}_i) 
- \sum_i y^{r,e}_i \log(p^{r,e}_i)
\end{equation}
where $y^{r,s}, y^{r,e} \in \mathbb{R}^{(n+2)}$ are true start and end positions for OTs given a specific AT.

In addition, for the right part, we also obtain the sentiment polarity 
\begin{equation}
    p^{r}_{cls} = softmax(W^{r}_{cls}h^{r}_{cls}+b^{r}_{cls})
\end{equation}
The cross entropy loss for the classification is 
\begin{equation}
    \mathcal{J}_{SC} = CE(p^{r}_{cls}, y_{cls})
\end{equation}
where $y_{cls}\in \mathbb{R}^{3}$ represents the true labels for sentiment polarities.
Then we want to minimize the final joint training loss 
\begin{equation}
    \mathcal{J} = \alpha \cdot \mathcal{J}_{AE} + \beta \cdot \mathcal{J}_{SC}+ \gamma \cdot \mathcal{J}_{AOE} \label{eqn_loss}
\end{equation}
where $\alpha,\beta, \gamma \in [0, 1]$ are hyper-parameters to control the contributions of objectives.

\subsection{MRC Dataset Conversion}
\begin{figure}
    \centering
    \includegraphics[width=0.47\textwidth]{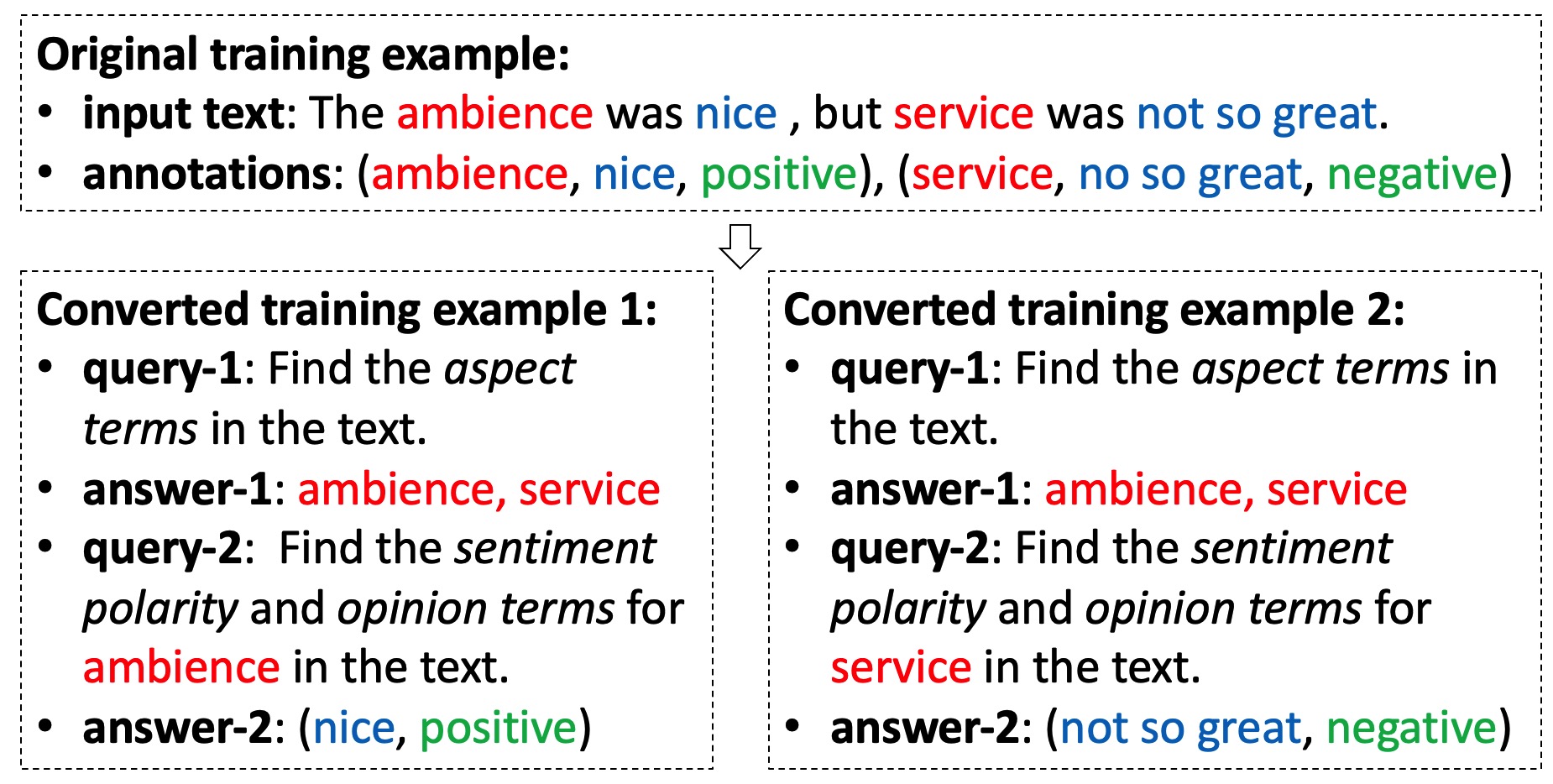}
    \caption{Dataset conversion.} \label{fig_convert}
\end{figure}

As illustrated in Figure \ref{fig_convert}, the original triple annotations have to be converted before it is fed into
the joint training dual-MRC model. Both MRCs use the input sentence as their contexts.
The left MRC is constructed with the query 
\begin{equation}
    q_1 = \text{``Find the aspect terms in the text."} \label{eqn_query_1}
\end{equation}
Then the answer to the left MRC is all ATs from the text.  
Given an AT, the right MRC is constructed with the query
\begin{eqnarray}
    q_2(AT) = \text{``Find the sentiment polarity and} \nonumber \\ 
    \text{opinion terms for AT in the text."} \label{eqn_query_2}
\end{eqnarray}
The output to the right MRC is all OTs and the sentiment polarity with respect to the given AT. 
An important problem is that number of right MRCs equals the number of ATs, therefore, the left MRC is \emph{repeated} for that number of times. 

\subsection{Inference Process}
For \emph{Triple}, we want to point out some differences between the training process and inference process.
During the training process, the ground truth of all ATs are known, then the right MRC can be constructed based on these ATs. 
Thus, the training process is end-to-end.
However, during the inference process, the ATs are the output of the left MRC.
Therefore, we inference the two MRCs in a pipeline, as in Algorithm \ref{algo}.
\begin{algorithm}
    \SetAlgoLined
    \KwIn{sentence $x$}
    \KwOut{$T=\{(a,o,s)\}$ triples }
    Initialize $T=\{\}$\;
    Input $x$ with the query $q_1$ described in (\ref{eqn_query_1}) as the left MRC, and output the AT candidates $A$\;
    If $A = \{\}$, return $T$\;
    \For{$a_i \in A$}
    {
        Input $x$ with the query $q_2$ described in (\ref{eqn_query_2}) as the right MRC, 
        and output the sentiment polarity $s$ and OTs $\{o_j, j= 1,2,...\}$\;
        $T \leftarrow T \bigcup \{(a_i, o_j, s),j=1,2,...\}$
    }
    Return $T$.
     \caption{The inference Process for Triple Extraction of the Dual-MRC Framework} \label{algo}
\end{algorithm}

The inference process of other tasks are similar. The task \emph{AE} uses the span output from the left MRC. 
\emph{AOE} and \emph{SC} use the span and classification outputs from the right MRC. 
\emph{AESC} and \emph{Pair} use a combination of them. 
Please refer to Table \ref{subtasks} for details.

\section{Experiments}
\subsection{Datasets}
Original datasets are from the Semeval Challenges\cite{pontiki-etal-2014-semeval, pontiki-etal-2015-semeval, pontiki-etal-2016-semeval}, where ATs and corresponding sentiment polarities are labeled. 
We evaluate our framework on three public datasets derived from them.

The first dataset is from \cite{DBLP:conf/aaai/WangPDX17}, where labels for opinion terms are annotated. All datasets share a fixed training/test split. 
 The second dataset is from \cite{fan2019target}, where (AT, OT) pairs are labeled. 
The third dataset is from \cite{peng2020knowing} where (AT, OT, SP) triples are labeled. 
A small number of samples with overlapping ATs and OTs are corrected. 
Also, $20\%$ of the data from the training set are randomly selected as the validation set. 
For detailed statistics of the datasets, please refer to the original papers.

\begin{table*}
    \centering
    \begin{tabular}{c|cccc|cccc|cccc}
    \hline
             & \multicolumn{4}{c|}{14res}         & \multicolumn{4}{c|}{14lap}         & \multicolumn{4}{c}{15res}          \\ \cline{2-13} 
             & AE    & OE    & SC    & AESC       & AE    & OE    & SC    & AESC       & AE    & OE    & SC    & AESC       \\ \hline
    SPAN-BERT    & \textbf{86.71} & -     & 71.75 & 73.68      & 82.34 & -     & 62.50 & 61.25      & 74.63 & -     & 50.28 & 62.29      \\
    IMN-BERT      & 84.06 & 85.10 & 75.67 & 70.72      & 77.55 & 81.00 & 75.56 & 61.73      & 69.90 & 73.29 & 70.10 & 60.22      \\
    RACL-BERT     & 86.38 & 87.18 & 81.61 & 75.42      & 81.79 & 79.72 & 73.91 & 63.40      & 73.99 & 76.00 & \textbf{74.91} & \textbf{66.05}      \\ \hline
    Dual-MRC & 86.60 & -     & \textbf{82.04}     & \textbf{75.95} & \textbf{82.51}     & -     & \textbf{75.97}     & \textbf{65.94} & \textbf{75.08}     & -     & 73.59    & 65.08 \\ \hline
    \end{tabular} 
    \caption{Results for \emph{AE}, \emph{SC} and \emph{AESC} on the datasets annotated by \cite{DBLP:conf/aaai/WangPDX17}. 
    \emph{OE} is not applicable to our proposed framework. All tasks are evaluated with F1. 
    Baseline results are directly taken from \cite{chen-qian-2020-relation}. Our model is based on BERT-Large-Uncased. 
    $20\%$ of the data from the training set are randomly selected as the validation set. The results are the average scores of 5 runs with random initialization.}
    \label{exp_1}
\end{table*} 
\begin{table*}
    \centering
    \begin{tabular}{c|ccc|ccc|ccc|ccc}
    \hline
                & \multicolumn{3}{c|}{14res}     & \multicolumn{3}{c|}{14lap}     & \multicolumn{3}{c|}{15res}     & \multicolumn{3}{c}{16res}      \\ \cline{2-13} 
                & P     & R     & F1             & P     & R     & F1             & P     & R     & F1             & P     & R     & F1             \\ \hline
    IOG       & 82.38 & 78.25 & 80.23          & 73.43 & 68.74 & 70.99          & 72.19 & 71.76 & 71.91          & 84.36 & 79.08 & 81.60          \\
    LOTN      & 84.00 & \textbf{80.52} & 82.21          & 77.08 & 67.62 & 72.02          & 76.61 & 70.29 & 73.29          & \textbf{86.57} & \textbf{80.89} & \textbf{83.62} \\ \hline
    Dual-MRC  & \textbf{89.79} & 78.43 & \textbf{83.73} & \textbf{78.21} & \textbf{81.66} & \textbf{79.90} & \textbf{77.19} & \textbf{71.98} & \textbf{74.50} & 86.07 & 80.77 & 83.33          \\ \hline
    \end{tabular} 
    \caption{Results for \emph{AOE} on the datasets annotated by \cite{fan2019target}. 
    Baseline results are directly taken from \cite{DBLP:conf/aaai/WuZDHC20}. Our model is based on BERT-Base-Uncased.}
    \label{exp_2}
\end{table*}

\subsection{Subtasks and Baselines}
There exist three research lines in ABSA where each research line with different data annotations, ABSA substasks, baselines and experimental settings. 
To fairly compare our proposed framework with previous baselines, we should specify them clearly for each research line.

Using the dataset from \cite{DBLP:conf/aaai/WangPDX17}, the following baselines were evaluated for \emph{AE}, \emph{OE}, \emph{SC} and \emph{AESC}:
\begin{itemize}
    \item \textbf{SPAN-BERT} \cite{hu-etal-2019-open} is a pipeline method for \emph{AESC} which takes BERT as the backbone network. A span boundary detection module is  used for \emph{AE},  then followed by a polarity classifier based on span representations for \emph{SC}.
    \item \textbf{IMN-BERT} \cite{he-etal-2019-interactive} is an extension  of IMN \cite{he-etal-2019-interactive} with BERT as the backbone. IMN is a multi-task learning method involving joint training for \emph{AE} and \emph{SC}. A message-passing architecture is introduced in IMN to boost the performance of \emph{AESC}.
    \item \textbf{RACL-BERT} \cite{chen-qian-2020-relation}  is a stacked multi-layer network based on BERT encoder and is the state of the art method for \emph{AESC}. A Relation propagation mechanism is utilized in RACL to capture the interactions between subtasks (i.e. \emph{AE}, \emph{OE}, \emph{SC}).
\end{itemize}
Using the dataset from \cite{fan2019target}, the following baselines were evaluated for \emph{AOE}:
\begin{itemize}
    \item \textbf{IOG} \cite{fan2019target} is the first model proposed to address \emph{AOE}, which adopts six different BLSTMs to extract corresponding opinion terms for aspects given in advance. 
    \item \textbf{LOTN} \cite{DBLP:conf/aaai/WuZDHC20} is the state of the art method for AOE, which transfer latent opinion information from external  sentiment classification datasets to improve the performance. 
\end{itemize}
Using the dataset from \cite{peng2020knowing}, the following baselines were evaluated for \emph{AESC}, \emph{Pair} and \emph{Triple}:
\begin{itemize}
    \item \textbf{RINANTE} \cite{dai-song-2019-neural} is a weakly supervised co-extraction method for \emph{AE} and \emph{OE} which make use of the dependency relations of words in a sentence. 
    \item \textbf{CMLA} \cite{DBLP:conf/aaai/WangPDX17} is a multilayer attention network for \emph{AE} and \emph{OE}, where each layer consists of a couple of attentions with tensor operators.
    \item \textbf{Li-unified-R} \cite{peng2020knowing} is a modified variant of Li-unified\cite{li2019unified}, which is originally for \emph{AESC} via a unified tagging scheme.  Li-unified-R only adapts the original \emph{OE} module for opinion term extraction.
    \item \textbf{Peng-two-stage} \cite{peng2020knowing} is a two-stage framework with separate models for different subtasks in ABSA and is the state-of-the-art method for \emph{Triple}.
\end{itemize}

\begin{table*}
    \centering
    \small 
    \begin{tabular}{cc|ccc|ccc|ccc|ccc}
    \hline
    \multicolumn{1}{l}{}                         &                & \multicolumn{3}{c|}{14res}     & \multicolumn{3}{c|}{14lap}     & \multicolumn{3}{c|}{15res}     & \multicolumn{3}{c}{16res}      \\ \cline{3-14} 
    \multicolumn{1}{l}{}                         &                & P     & R     & F1             & P     & R     & F1             & P     & R     & F1             & P     & R     & F1             \\ \hline
    \multicolumn{1}{c|}{\multirow{5}{*}{AESC}}   & RINANTE        & 48.97 & 47.36 & 48.15          & 41.20 & 33.20 & 36.70          & 46.20 & 37.40 & 41.30          & 49.40 & 36.70 & 42.10          \\
    \multicolumn{1}{c|}{}                        & CMLA           & 67.80 & 73.69 & 70.62          & 54.70 & 59.20 & 56.90          & 49.90 & 58.00 & 53.60          & 58.90 & 63.60 & 61.20          \\
    \multicolumn{1}{c|}{}                        & Li-unified-R   & 73.15 & 74.44 & 73.79          & 66.28 & 60.71 & 63.38          & 64.95 & \textbf{64.95} & 64.95          & 66.33 & \textbf{74.55} & 70.20          \\
    \multicolumn{1}{c|}{}                        & Peng-two-stage & 74.41 & 73.97 & 74.19          & 63.15 & 61.55 & 62.34          & \textbf{67.65} & 64.02 & \textbf{65.79} & \textbf{71.18} & 72.30 & \textbf{71.73} \\ \cline{2-14}
    \multicolumn{1}{c|}{}                        & Dual-MRC       & \textbf{76.84} & \textbf{76.31} &\textbf{76.57}  & \textbf{67.45} & \textbf{61.96} & \textbf{64.59} & 66.84 & 63.52 & 65.14          & 69.18 & 72.59 & 70.84          \\ \hline
    \multicolumn{1}{c|}{\multirow{5}{*}{Pair}}   & RINANTE        & 42.32 & 51.08 & 46.29          & 34.40 & 26.20 & 29.70          & 37.10 & 33.90 & 35.40          & 35.70 & 27.00 & 30.70          \\
    \multicolumn{1}{c|}{}                        & CMLA           & 45.17 & 53.42 & 48.95          & 42.10 & 46.30 & 44.10          & 42.70 & 46.70 & 44.60          & 52.50 & 47.90 & 50.00          \\
    \multicolumn{1}{c|}{}                        & Li-unified-R   & 44.37 & 73.67 & 55.34          & 52.29 & 52.94 & 52.56          & 52.75 & 61.75 & 56.85          & 46.11 & 64.55 & 53.75          \\
    \multicolumn{1}{c|}{}                        & Peng-two-stage & 47.76 & 68.10 & 56.10          & 50.00 & 58.47 & 53.85          & 49.22 & \textbf{65.70} & 56.23          & 52.35 & 70.50 & 60.04          \\ \cline{2-14}
    \multicolumn{1}{c|}{}                        & Dual-MRC       & \textbf{76.23} & \textbf{73.67} & \textbf{74.93} & \textbf{65.43} & \textbf{61.43} & \textbf{63.37} & \textbf{72.43} & 58.90 & \textbf{64.97} & \textbf{77.06} & \textbf{74.41} & \textbf{75.71} \\ \hline
    \multicolumn{1}{c|}{\multirow{5}{*}{Triple}} & RINANTE        & 31.07 & 37.63 & 34.03          & 23.10 & 17.60 & 20.00          & 29.40 & 26.90 & 28.00          & 27.10 & 20.50 & 23.30          \\
    \multicolumn{1}{c|}{}                        & CMLA           & 40.11 & 46.63 & 43.12          & 31.40 & 34.60 & 32.90          & 34.40 & 37.60 & 35.90          & 43.60 & 39.80 & 41.60          \\
    \multicolumn{1}{c|}{}                        & Li-unified-R   & 41.44 & 68.79 & 51.68          & 42.25 & 42.78 & 42.47          & 43.34 & 50.73 & 46.69          & 38.19 & 53.47 & 44.51          \\
    \multicolumn{1}{c|}{}                        & Peng-two-stage & 44.18 & 62.99 & 51.89          & 40.40 & 47.24 & 43.50          & 40.97 & \textbf{54.68} & 46.79          & 46.76 & 62.97 & 53.62          \\ \cline{2-14}
    \multicolumn{1}{c|}{}                        & Dual-MRC       & \textbf{71.55} & \textbf{69.14} &\textbf{70.32}  & \textbf{57.39} & \textbf{53.88} & \textbf{55.58} & \textbf{63.78} & 51.87 & \textbf{57.21} & \textbf{68.60} & \textbf{66.24} & \textbf{67.40} \\ \hline
    \end{tabular} 
    \caption{Results for \emph{AESC}, \emph{Pair} and \emph{Triple} on the datasets annotated by \cite{peng2020knowing}. 
    Baseline results are directly taken from \cite{peng2020knowing}. Our model is based on BERT-Base-Uncased.}
    \label{exp_3}
    \end{table*}

\subsection{Model Settings}
We use the BERT-Base-Uncased\footnote{https://github.com/google-research/bert} or BERT-Large-Uncased
as backbone models for our proposed model depending on the baselines. 
Please refer to \cite{devlin2019bert} for model details of BERT.
We use Adam optimizer with a learning rate of $2e^{-5}$ and warm up over the first $10\%$ steps to train for 3 epochs. 
The batch size is $12$ and a dropout probability of $0.1$ is used. 
The hyperparameters $\alpha, \beta, \gamma$ for the final joint training loss in Equation \ref{eqn_loss} are not sensitive to results, so we fix them as $1/3$ in our experiments.
The logit thresholds of heuristic multi-span decoding algorithms \cite{hu-etal-2019-open} are very sensitive to results and they are manually tuned on each dataset, and other hyperparameters are kept default. All experiments are conducted on a single Tesla-V100 GPU.

\subsection{Evaluation Metrics}
For all tasks in our experiments, we use the precision (P), recall (R), and F1 scores
\footnote{We use F1 as the metric for aspect-level sentiment classification following \cite{chen-qian-2020-relation}} as evaluation metrics since
a predicted term is correct if it exactly matches a gold term.

\subsection{Main Results}
As mentioned previously, there are three research lines with different datasets, ABSA substasks, baselines and experimental settings. 
For each research line, we keep the same dataset and experimental setting, and compare our proposed dual-MRC framework with the baselines and present our results in Table \ref{exp_1}, Table \ref{exp_2} and Table \ref{exp_3}.

First, we compare our proposed method for \emph{AE}, \emph{SC} and \emph{AESC} on the dataset from \cite{DBLP:conf/aaai/WangPDX17}. 
\emph{OE} is not applicable to our proposed framework
\footnote{If needed, we can train a separate model with the query ``\emph{Find the opinion terms in the text.}'' for \emph{OE}.}. 
Since the pair-wise relations of (AT, OT) are not annotated in this dataset, we use the right part of our model for classification only. 
$20\%$ of the data from the training set are randomly selected as the validation set. 
The results are the average scores of 5 runs with random initialization and they are shown in Table \ref{exp_1}. 
We adopt BERT-Large-Uncased as our backbone model since the baselines use it too.
All the baselines are BERT based and our results achieve the first or second place comparing to them. 
Recall that our approach is inspired by SPAN-BERT, which is a strong baseline for extraction tasks.
Our results are close to SPAN-BERT in \emph{AE}.
However, with the help of MRC, we achieve much better results in \emph{SC} and \emph{AESC}. 

Second, we compare our proposed method for \emph{AOE} on the dataset from \cite{fan2019target}, where the pair-wise (AT, OT) relations are annotated. 
This task can be viewed as a trivial case of our proposed full model. The results are shown in Table \ref{exp_2}. 
BERT-Base-Uncased is used as our backbone model. 
Although  the result for 16res is a little bit lower than LOTN, most of our results significantly outperform the previous baselines. 
It indicates our model has advantage in matching AT and OT. In particular, our model performs much better than baselines on lap14. It is probably due to 
the domain difference between the laptop (14lap) comments and the restaurant comments (14res/15res/16res). 

Third, we compare our proposed method for \emph{AESC}, \emph{Pair} and \emph{Triple} on the dataset from \cite{peng2020knowing}. 
The full model of our proposed framework is implemented.  The results are shown in Table \ref{exp_3}. 
BERT-Base-Uncased is used as our backbone model. 
Our results significantly outperform the baselines, especially in the precision scores of extraction the pair-wise (AT, OT) relations. 
Note that Li-unified-R and Peng-two-stage both use the unified tagging schema. For extraction tasks, span based methods outperform the unified tagging schema for extracting terms, probably because determining the start/end positions is easier than determining the label for every token. 
More precisely, for the unified tagging schema, there are at 7 possible choices for each token, say $\{$B-POS, B-NEU, B-NEG, I-POS, I-NEU, I-NEG, O$\}$, so there are $7^n$ total choices. 
For span based methods, there are at 4 possible choices for each token, say $\{$IS-START, NOT-START, IS-END, NOT-END$\}$, then there are $4^n (\ll 7^n)$ total choices. 
Our proposed method combines MRC and span based extraction, and it has huge improvements for \emph{Pair} and \emph{Triple}. 

\begin{table*}
    \centering
    \small
    \begin{tabular}{l|l|l|l|l|l}
    \hline
    Example                                                                                                                                                           & Ground Truth                                                                                                                                    & Our model                                                                                                                                                                              & Peng-two-stage         \\ \hline
    \begin{tabular}[c]{@{}l@{}}Rice is too dry, tuna \\was n’t so fresh either.\end{tabular}                                                                    & \begin{tabular}[c]{@{}l@{}}(Rice, too dry, NEG) \\ (tuna, was n’t so fresh, NEG)\end{tabular}                                                      & \begin{tabular}[c]{@{}l@{}}(Rice, too dry, NEG) \\ (tuna, was n’t so fresh, NEG)\end{tabular}                                                                                          & \begin{tabular}[c]{@{}l@{}}(Rice, too dry, NEG) \\ (tuna, was n’t so fresh, NEG), \\ (Rice, was n’t so fresh, NEG)\XSolidBrush \\ (tuna, too dry, NEG)\XSolidBrush\end{tabular}              \\ \hline
    \begin{tabular}[c]{@{}l@{}}I am pleased with \\ the fast log on, speedy \\ WiFi connection and \\ the long battery life.\end{tabular}                             & \begin{tabular}[c]{@{}l@{}}(log on, pleased, POS) \\ (log on, fast, POS) \\ (WiFi connection, speedy, POS) \\ (battery life, long, POS)\end{tabular} & \begin{tabular}[c]{@{}l@{}}(log on, pleased, POS) \\ (log on, fast, POS) \\ (WiFi connection, speedy, POS)\\  (WiFi connection, pleased, POS)\XSolidBrush, \\ (battery life, long, POS)\end{tabular} & \begin{tabular}[c]{@{}l@{}}(log, pleased, POS)\XSolidBrush, \\ (log, fast, POS)\XSolidBrush \\ (WiFi connection, speedy, POS) \\ (battery life, long, POS)\end{tabular}          \\ \hline
    \begin{tabular}[c]{@{}l@{}}The service was  \\ exceptional - sometime  \\there was a feeling that \\ we were served by the \\ army of friendly waiters.\end{tabular} & \begin{tabular}[c]{@{}l@{}}(service, exceptional, POS) \\ (waiters, friendly, POS)\end{tabular}                                                    & \begin{tabular}[c]{@{}l@{}}(service, exceptional, POS) \\ (waiters, friendly, POS)\end{tabular}                                                                                        & \begin{tabular}[c]{@{}l@{}}(service, exceptional, POS) \\ (waiters, friendly, POS)\end{tabular}                                                                        \\ \hline
    \end{tabular}
    \caption{Case study of task \emph{Triple}. Wrong predictions are marked with \XSolidBrush. The three examples are extractly the same as the ones selected by \cite{peng2020knowing}.}
    \label{case_study}
\end{table*}

\subsection{Analysis on Joint Learning}
We give some analysis on the effectiveness of joint learning. 
The experimental results on the dataset from \cite{peng2020knowing} are shown in Table \ref{analysis_joint}.
Overall, from the experimental results, adding one or two learning objectives does not affect much in F-1 scores. 
However, joint learning is more efficient and it can handle more tasks with one single model.  

\begin{table}
    \centering
    \small
    \begin{tabular}{c|c|c|c|l|l|l|l}
        \hline
        \multirow{2}{*}{Task}                   & Left  & \multicolumn{2}{c|}{Right} & \multirow{2}{*}{14res} & \multirow{2}{*}{14lap} & \multirow{2}{*}{15res} & \multirow{2}{*}{16res} \\ \cline{2-4}
                              & e & c      & e      &                        &                        &                        &                        \\ \hline
        \multirow{2}{*}{AESC} & $\surd$        & $\surd$            &                & 76.31                  & 63.95                  & 65.43                  & 69.48                  \\ \cline{2-8} 
                              & $\surd$        & $\surd$            & $\surd$             & 76.57           & 64.59           & 65.14         & 70.84         \\ \hline
        \multirow{2}{*}{Pair} & $\surd$        &               & $\surd$             & 76.33                  & 65.26                  & 65.21                  & 76.61                  \\ \cline{2-8} 
                              & $\surd$        & $\surd$            & $\surd$             & 74.93           & 63.37          & 64.97          & 75.71           \\ \hline
        \multirow{2}{*}{AE}   & $\surd$        &               &                & 82.80                  & 78.35                  & 78.22                  & 82.16                  \\ \cline{2-8} 
                              & $\surd$        & $\surd$            & $\surd$             & 82.93           &      77.31                   & 76.08       & 81.20          \\ \hline
        \end{tabular}
    \caption{Results on the analysis of joint learning for \emph{AESC} and \emph{Pair} on the dataset from \cite{peng2020knowing}.
    In the table, the letter e stands for extraction and the letter c stands for classification. } \label{analysis_joint}
    \end{table}
    
For the task \emph{AESC}, we compare the results with or without the span based extraction output from the right part of our model.
By jointly learning to extract the opinion terms for a given aspect, the result of aspect-level sentiment classification is improved a little bit.  
It makes sense because extracted OTs are useful for identifying the sentiment polarity of the given AT. 

For the task \emph{Pair}, we compare the results with or without the classification output from the right part of our model. 
The F-1 scores for OT extraction decrease a little bit when the sentiment classification objective is added. 
The reason might be that the sentiment polarity can point to multiple OTs in a sentence where some OTs are not paired with the given AT.

\subsection{Case Study}
To validate the effectiveness of our model, we compare our method based on exactly the same three examples in the baseline \cite{peng2020knowing} as its source code is not public. 
The results are shown in Table \ref{case_study}.

The first example shows our MRC based approach performs better in matching AT and OT. 
Peng's approach matches ``\emph{tuna}'' and ``\emph{too dry}'' by mistake while our approach converts the matching problem to a MRC problem. 
The second example shows the span based extraction method is good at detecting boundaries of entities. 
Our approach successfully detects ``\emph{log on}'' while Peng's approach detects ``\emph{log}'' by mistake.
Moreover, the sentiment classification result indicates that our MRC based approach is also good at \emph{SC}.

\begin{figure}[h!]
    \centering
    \includegraphics[width=0.45\textwidth]{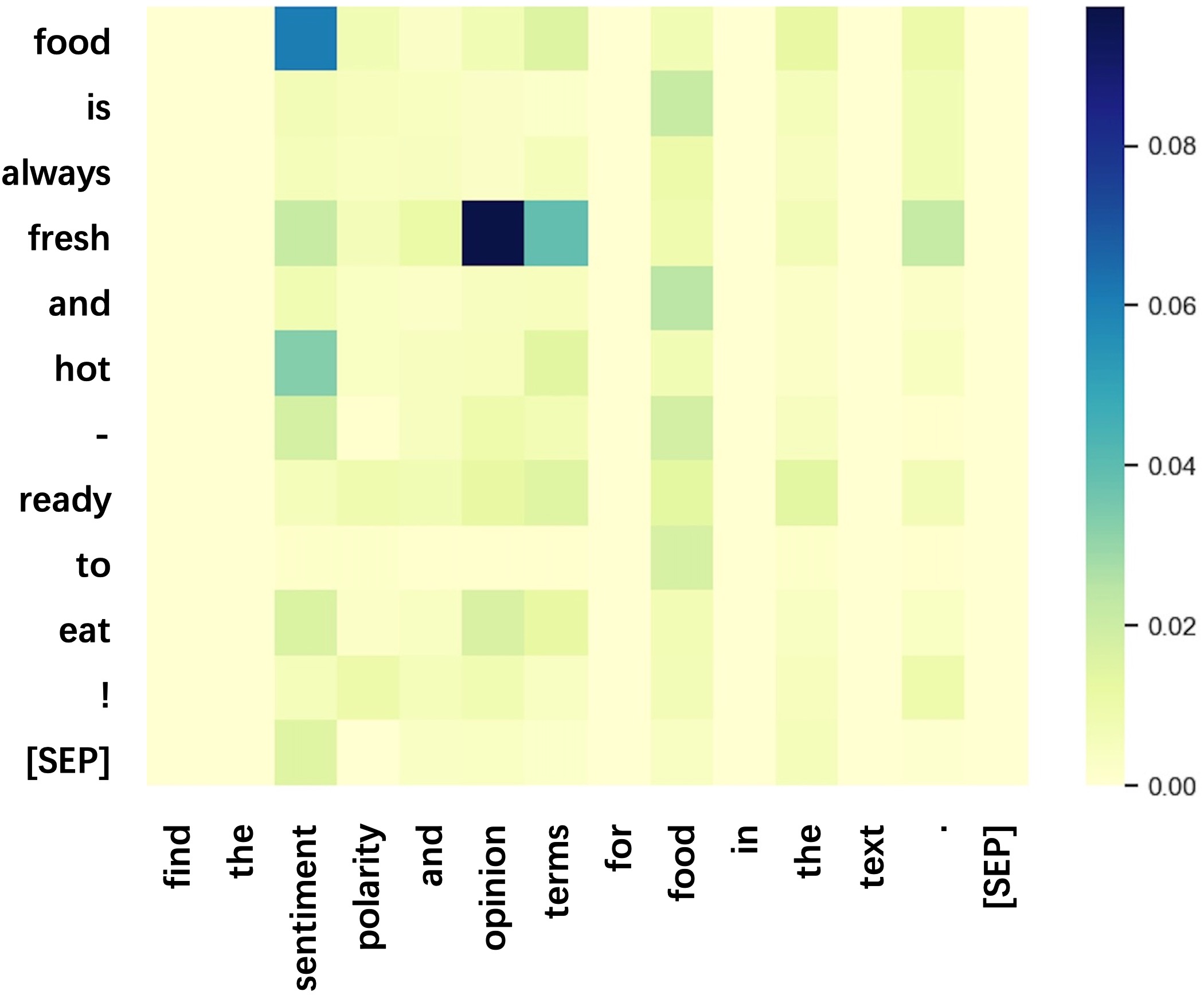}
    \caption{An example of attention matrices for the input text and query.} \label{fig_att}
\end{figure}

We plot in Figure \ref{fig_att} the attention matrices from our fine-tuned model between the input text and the query.
As we can see, the ``\emph{opinion term}'' has high attention scores with ``\emph{fresh}'', 
and ``\emph{sentiment}'' has high attention scores with ``\emph{food/fresh/hot}''. 
As a result, the queries can capture important information for the task via self-attentions. 

\section{Conclusions}
In this paper, we propose a joint training dual-MRC framework to handle all ABSA subtasks of aspect based sentiment analysis (ABSA)
in one shot, where the left MRC is for aspect term extraction and the right MRC is for aspect-oriented opinion term extraction and sentiment classification.
The original dataset is converted and fed into dual-MRC to train jointly. For three research lines, 
experiments are conducted and are compared with different ABSA subtasks and baselines. 
Experimental results indicate that our proposed framework outperforms all compared baselines.

\bibliography{dual_mrc.bib}
\end{document}